\documentclass[]{fairmeta}
\usepackage{xcolor}
\usepackage[utf8]{inputenc}
\usepackage{multirow}
\usepackage{nameref}

\usepackage[utf8]{inputenc} % allow utf-8 input
\usepackage[T1]{fontenc}    % use 8-bit T1 fonts
\usepackage{url}            % simple URL typesetting
\usepackage{amsfonts}       % blackboard math symbols
\usepackage{nicefrac}       % compact symbols for 1/2, etc.
\usepackage{microtype}      % microtypography
\usepackage{xcolor}         % colors
\usepackage{wrapfig}        % wrapfigure
\usepackage[nohyperlinks]{acronym} 
\usepackage{bbm}            % for binary indicator function
\usepackage{xspace}         % eg/etal onedot definition 
\usepackage{enumitem}       % bullet point in intro
\usepackage{bm}             % bold math symbols
\usepackage{multirow}

\usepackage[textsize=tiny]{todonotes}

\usepackage{color}

\usepackage{makecell} % newline in a table cell
\usepackage{fontawesome5}  % for warning icon
\usepackage{amssymb}
\usepackage{adjustbox}
\usepackage{amsthm}
\usepackage{amsmath}        % For theorems and declare math operator
\usepackage{listings}       % to render the code nicely
\usepackage{mathtools}      % for \coloneqq

\usepackage{booktabs}       % professional-quality tables
\usepackage{nicematrix}
\usepackage{bm}             % bold math 

\makeatletter
\DeclareRobustCommand\onedot{\futurelet\@let@token\@onedot}
\def\@onedot{\ifx\@let@token.\else.\null\fi\xspace}

\makeatother

\definecolor{safe}{HTML}{69A831}   % green
\definecolor{unsafe}{HTML}{A657D6} % purple
\definecolor{methodfg}{HTML}{FF8800}
\definecolor{methodbg}{RGB}{255,244,227}
\definecolor{lightgray}{HTML}{f0f0f0}
\definecolor{mypurple}{HTML}{A13FDB} % sft methods
\definecolor{mypink}{HTML}{E14B9E} % other methods
\definecolor{myblue}{HTML}{456AFF} % trigger words
\definecolor{myred}{HTML}{F20D0D} % unsafe
\definecolor{mygreen}{HTML}{4ED65C} % safe green

\acrodef{lrm}[LRM]{large reasoning model}
\acrodef{cot}[CoT]{chain-of-thought}
\acrodef{sft}[SFT]{supervised finetuning}
\acrodef{rlhf}[RLHF]{reinforcement learning from human feedback}
\acrodef{dapo}[DAPO]{decouple clip and dynamic sampling policy optimization}
\acrodef{grpo}[GRPO]{group relative policy optimization }
\acrodef{rl}[RL]{reinforcement learning}
\acrodef{rlvr}[RLVR]{Reinforcement Learning with Verifiable Rewards}
\acrodef{ipr}[IPR]{iterative prefill reset}
\acrodef{mdp}[MDP]{markov decision process}
\acrodef{llm}[LLM]{large language model}

\begin{document}
\title{Skills-Coach: A Self-Evolving Skill Optimizer via Training-Free GRPO}

\author[1, *, \ddagger]{Yu Tian}
\author[2,3, *]{Jiawei Chen}
\author[4]{Lifan Zheng}
\author[5]{Mingxiang Tao}
\author[6]{Xinyi Zeng}
\author[2]{Zhaoxia Yin}
\author[6]{Hang Su}
\author[1, \dagger]{Xian Sun}

\affiliation[1]{University of Chinese Academy of Sciences}
\affiliation[2]{East China Normal University}
\affiliation[3]{Zhongguancun Academy}
\affiliation[4]{Southeast University}
\affiliation[5]{Hainan University}
\affiliation[6]{Tsinghua University}
\contribution[\dagger]{Corresponding author}
\contribution[*]{Equal contribution}
\contribution[\ddagger]{Project Leader}

\abstract{We introduce Skills-Coach, a novel automated framework designed to significantly enhance the self-evolution of skills within Large Language Model (LLM)-based agents. Addressing the current fragmentation of the skill ecosystem, Skills-Coach explores the boundaries of skill capabilities, thereby facilitating the comprehensive competency coverage essential for intelligent applications. The framework comprises four core modules: a Diverse Task Generation Module that systematically creates a comprehensive test suite for various skills; a Lightweight Optimization Module dedicated to optimizing skill prompts and their corresponding code; a Comparative Execution Module facilitating the execution and evaluation of both original and optimized skills; and a Traceable Evaluation Module, which rigorously evaluates performance against specified criteria. Skills-Coach offers flexible execution options through its virtual and real modes. To validate its efficacy, we introduce Skill-X, a comprehensive benchmark dataset consisting of 48 diverse skills. Experimental results demonstrate that Skills-Coach achieves significant performance improvements in skill capability across a wide range of categories, highlighting its potential to advance the development of more robust and adaptable LLM-based agents.
}

\date{\today}
\correspondence{Yu Tian \email{tianyu181@mails.ucas.ac.cn}, Xian Sun \email{sunxian@aircas.ac.cn}}

% You can add additional metadata fields as follows 
\metadata[Code]{\url{https://github.com/T1aNS1R/Skills-Coach}} 
\metadata[Clawhub]{\url{https://clawhub.ai/t1ans1r/skills-coach}}
% \metadata[Blogpost]{\url{https://ai.meta.com/blog/?page=1}}

\maketitle
\begin{figure*}[htbp] 
    \centering 
    \includegraphics[width=1\textwidth]{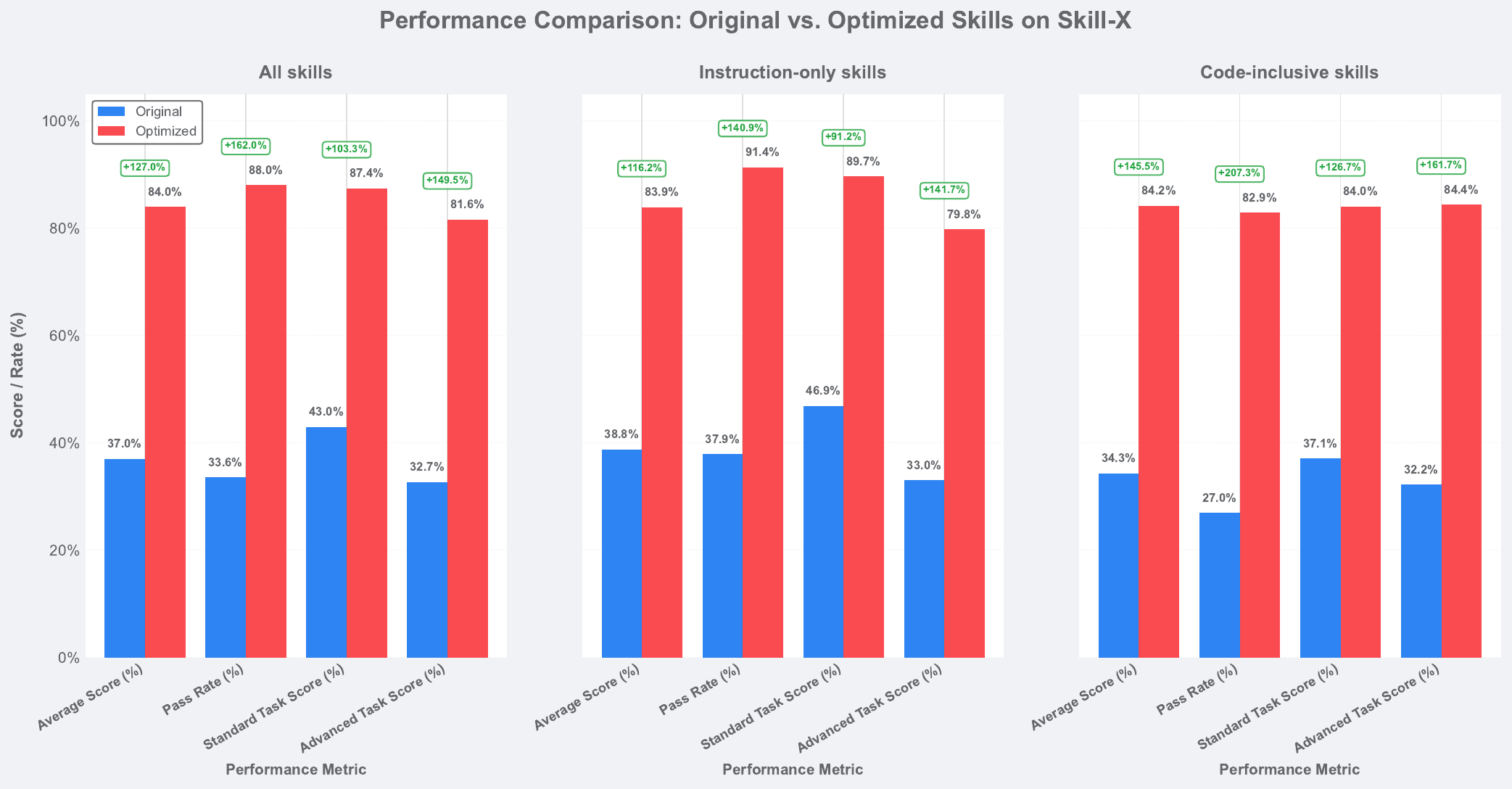}
    \caption{Performance Comparison of Original vs. Optimized Skills across Skill-X.}
    \label{fig:skill_performance_comparison} 
\end{figure*}
\section{Introduction}
\label{section: intro}

Driven by the rapid advancement of Large Language Model (LLM)-based agents\cite{yao2022react,schick2023toolformer,wang2023voyager}, skills, serving as a modular capability extension mechanism, are profoundly reshaping the deployment of intelligent applications across diverse industries\cite{hong2023metagpt,wu2024autogen}. Essentially, skills function as encapsulated modules comprising instructions, scripts, and resources,\footnote{https://github.com/anthropics/skills}\footnote{https://agentskills.io/home} enabling LLM-based agents to dynamically load specific capabilities and ensure precise execution in specialized tasks. Their applications span a broad spectrum, ranging from generating enterprise-specific brand documents and conducting organization-level data analysis to automating personal routines\cite{shen2023hugginggpt,xie2024osworld}. As agent technologies continue to flourish, developers across various domains have contributed a massive volume of skills tailored to their unique operational workflows and practical use cases; on the Clawhub platform alone, the repository has already exceeded 56,000 skills.\footnote{https://clawhub.ai/skills} 

However, the vast majority of these skills are developed by individuals to solve highly specific problems, meaning their design inherently focuses on localized use cases \cite{qin2024toolllm}. Consequently, they struggle to systematically cover the comprehensive functional requirements of complex, specialized tasks. This paradigm has engendered a skill ecosystem characterized by \textit{abundant volume but fragmented coverage}, leaving users tackling multifaceted tasks still grappling with functional gaps and integration bottlenecks\cite{patil2024gorilla,li2023api}. Such a fragmented landscape hinders the robust and scalable deployment of LLM-based agents, limiting their full potential.

Motivated by these limitations inherent in the current application of skills by LLM-based agents, we pose the following primary question: \textbf{Can an agent autonomously explore the capability boundaries of its existing skills and proactively expand them to achieve skill self-evolution?} We further decompose this question into three critical sub-questions:

\textbf{1) How can boundary-probing tasks be generated automatically?}
The prerequisite for exploring skill boundaries lies in constructing a challenging and comprehensive task set. If the tasks are too trivial, they fail to reach the upper limits of a skill's capabilities; if they lack systematicity, the exploration yields unrepresentative results. Therefore, the agent must be capable of autonomously generating test cases\cite{wang2023self,xu2023wizardlm}, systematically constructing boundary-testing samples through the induction and abstraction of existing tasks.

\textbf{2) How can skills achieve self-evolution?}
Beyond merely identifying capability boundaries, the agent should be able to transcend them by refining the skills themselves. This involves extracting patterns from failed exploration attempts and dynamically updating the skill modules. Concurrently, it must ensure that newly generated skills remain consistent and complementary with the existing skill ecosystem, thereby preventing capability degradation or conflicts.

\textbf{3) How can skill capabilities be effectively evaluated?}
Constructing a robust evaluation framework is essential to ensure the correct trajectory of skill evolution. An effective framework must precisely quantify a skill's performance across specific dimensions. This requires the evaluator to not only characterize the generalization boundaries and evolutionary potential of a skill but also account for its collaborative efficacy with the agent, providing reliable feedback to guide the self-evolution process.

To address the aforementioned challenges and assist developers in effectively utilizing and refining skills, we propose Skills-Coach, an automated framework for Skill Self-Evolution. Targeting a specific skill, Skills-Coach systematically explores its capability boundaries, conducts an in-depth analysis of potential optimization spaces, automatically generates improved versions, and compiles the findings into structured reports. Consequently, it achieves a closed-loop iterative refinement and autonomous evolution of skill capabilities without requiring continuous human intervention\cite{yuan2024self}. Specifically, Skills-Coach consists of four core components:

1) A \textbf{Diverse Task Generation Module} that analyzes the specifications of a target skill to construct a comprehensive test suite. This suite encompasses various tasks covering both standard use cases and advanced edge cases.

2) A \textbf{Lightweight Optimization Module} that leverages Training-free Group Relative Policy Optimization (GRPO) to iteratively refine a skill's instruction (e.g., Skill.md) and associated code files based on their performance across training tasks.

3) A \textbf{Comparative Execution Module} that runs both the original and optimized versions of the target skill against all test tasks, systematically capturing outputs and execution logs for subsequent evaluation.

4) A \textbf{Traceable Evaluation Module} designed to leverage multi-dimensional criteria for assessing two skill versions. It systematically calculates comprehensive metrics, performs comparative analysis, and informs data-driven retention decisions.

Notably, Skills-Coach supports two distinct optimization and execution configurations: a \textit{virtual mode} and a \textit{real mode}. In virtual mode, the system completely bypasses the actual execution of commands or scripts. Instead, it estimates task completion by verifying the presence of evaluation-criteria-related keywords (such as "error handling" and "save") within the skill instruction, combined with deterministic random numbers generated from a hash of the skill's content. Conversely, in real mode, the agent deploys the original or optimized skill in a practical environment. It then precisely evaluates whether the skill has fulfilled the task requirements by analyzing actual output files, execution logs, and error messages.

To rigorously evaluate the effectiveness of our proposed framework and provide a valuable resource for the research community, we introduce a novel benchmark dataset named \textbf{Skill-X}. This dataset comprises 48 widely used skills curated from platforms such as ClawHub, Anthropic, and SkillSh, ensuring broad diversity and high practical utility. Extensive empirical evaluations demonstrate that Skills-Coach delivers substantial performance enhancements across a diverse spectrum of skill categories, highlighting its potential to advance the development of more robust and adaptable LLM agents.

% The diverse task generation module analyzes the specification of a target skill to construct a comprehensive test suite, generating diverse tasks that encompass both standard use cases and challenging edge cases. The training-free GRPO-based training module applies Training-Free Group Relative Policy Optimization (GRPO) to iteratively refine a skill's prompt file (e.g., Skill.md) and its associated code files, guided by performance across the training task set. The multi-mode execution module executes both the original and optimized versions of the target skill against the full test suite, capturing outputs and execution logs for subsequent evaluation. The traceable evaluation module assesses both skill versions using SpecCheck criteria, computes comprehensive performance metrics, conducts comparative analysis, and ultimately makes a data-driven retention decision. It is worth noting that Skills-Coach supports two operational modes: a virtual mode and a real mode. In virtual mode, the system estimates task completion by examining whether the skill documentation contains keywords related to the evaluation criteria (e.g., "error handling" and "save"), combined with deterministic random values derived from a hash of the skill's content, without executing any commands or scripts. In real mode, the agent deploys the target or optimized skill in a live environment and rigorously assesses whether the skill has genuinely fulfilled the task requirements by examining actual output files, execution results, and error messages.

\section{Skills Coach}
\label{section: coach}

\begin{figure*}[htbp] 
    \centering 
    \includegraphics[width=1\textwidth]{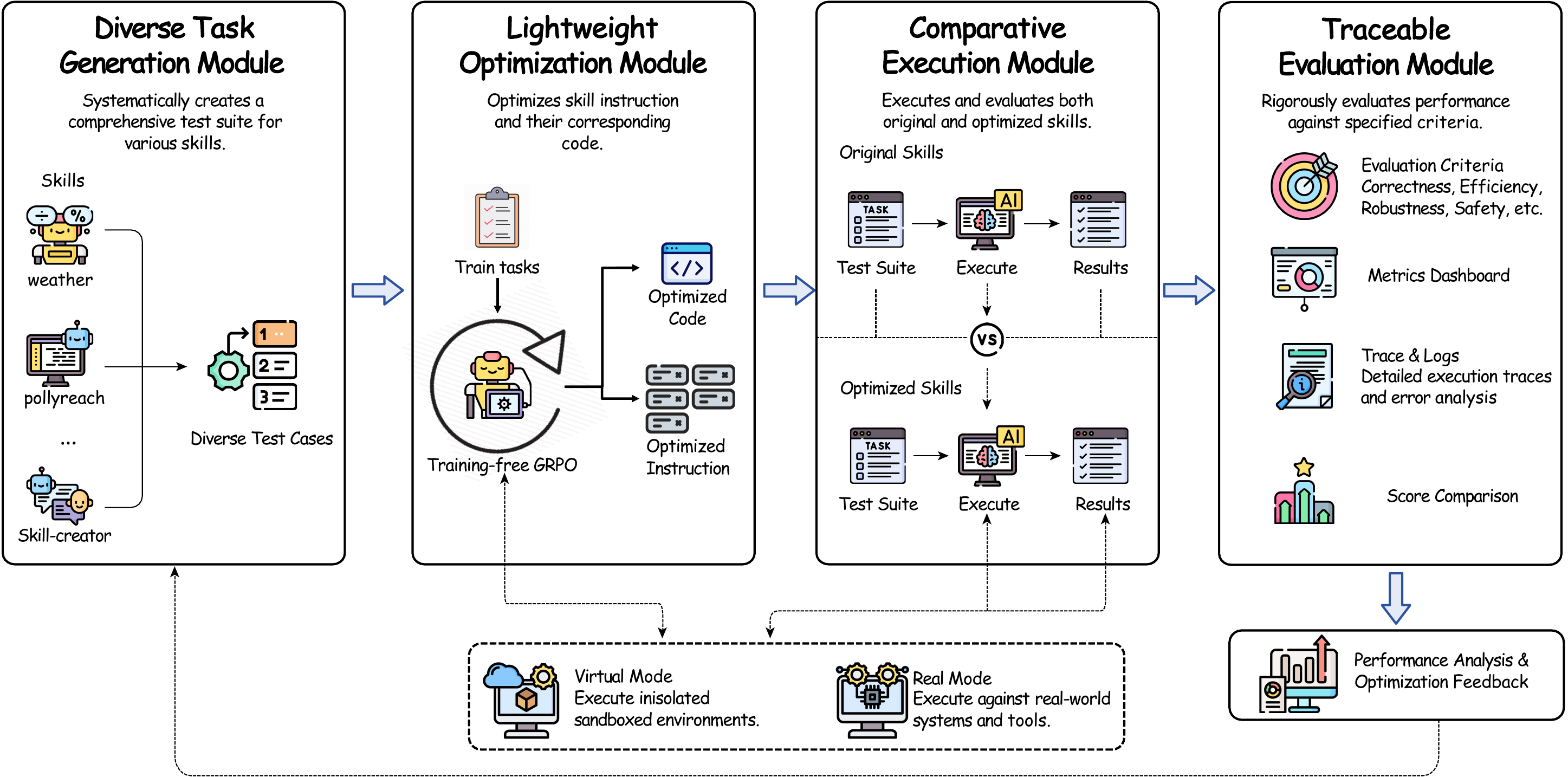}
    \caption{Overview of Skills-Coach. }
    \label{fig:overview} 
\end{figure*}

\subsection{Overview of Skills Coach}

To explore the capability boundaries of Skills and enable targeted optimization, we propose Skills Coach, an automated framework for Skill Self-Evolution. As shown in Figure \ref{fig:overview}, this framework is designed to comprehensively evaluate and optimize the execution capability and robustness of Skills through a structured pipeline. Skills Coach comprises four core modules: First, the Diverse Task Generation Module takes the Skill as input and automatically generates training and test sets covering standard use cases and challenging edge-case scenarios based on its functional instruction, thereby ensuring both comprehensive coverage and sufficient difficulty in evaluation. Second, the Lightweight Optimization Module performs targeted optimization of the target Skill at two levels (instruction and code) to effectively improve its performance. Subsequently, the Comparative Execution Module executes both the original Skill and the optimized Skill on the test set to obtain objective and comparable experimental results. Finally, the Traceable Evaluation Module systematically evaluates the execution results and produces detailed analytical reports, providing data-driven support for subsequent iterative optimization.

\subsection{Diverse Task Generation Module}

The Diverse Task Generation Module is crucial for Skills-Coach, constructing a comprehensive test suite that spans standard to complex edge-case scenarios by analyzing target skill specifications. This module generates diverse training and test sets essential for subsequent skill optimization and final performance evaluation. As all optimization decisions and performance metrics are derived from these tasks, their generation quality directly determines the accuracy and efficacy of the entire optimization pipeline. Consequently, the generated samples are designed with three core characteristics: 1) Sufficient diversity, avoiding the redundant accumulation of singular patterns; 2) Realistic boundary representation, reflecting the actual operational limits of the target skill rather than artificially idealized scenarios; and 3) Real-world applicability, ensuring that all tasks stem from practical demands to guarantee the empirical value of the optimization results.

To achieve these characteristics, the module comprehensively parses the skill's instruction files (e.g., Skill.md and Readme.md) to extract key information, including functional instructions, supported commands/parameters, I/O formats, constraints, and usage examples. Based on this data, the module categorizes the skill (executable or instruction-based), identifies runnable commands, and analyzes parameter roles and formatting. Employing regular expressions and structured parsing, it delineates the skill's capability boundaries, encompassing core functionalities, optional features, boundary conditions, and potential failure scenarios, thereby establishing a precise knowledge foundation for task generation.

A hierarchical generation strategy is employed, categorizing synthesized samples into three types: 1) \textit{Standard tasks}, covering routine operations like basic file processing; 2) \textit{Advanced tasks}, evaluating complex multi-step workflows and anomalous input handling; and 3) \textit{Boundary tasks}, probing operational limits through conditions such as min/max bounds, invalid inputs, and resource constraints. Notably, tasks within the test set maintain a difficulty level commensurate with the training set but feature entirely distinct contexts, rigorously assessing generalization capabilities over memorization. To ensure objectivity, all generated tasks are paired with automated validation criteria, utilizing metrics like output file existence, specific keyword inclusion, and regular expression compliance.

The module adheres to rigorous quality principles for test suite reliability, specifically: 1) \textit{Determinism}, ensuring reproducible results; 2) \textit{Strict Data Isolation}, separating training and test sets for true generalization assessment; 3) \textit{Diversity}, systematically varying input types, sizes, and formats; and 4) \textit{Objectivity}, using deterministically verifiable validation criteria. Furthermore, generated samples span eight evaluation dimensions—structural integrity, usability, example quality, technical depth, clarity, command coverage, error handling, and advanced scenarios. Each dimension includes at least six specific criteria, culminating in 51 discrete evaluation metrics (detailed in the Appendix \ref{appendix:1}). This comprehensive framework provides a robust foundation for optimization and performance evaluation.

\subsection{Lightweight Optimization Module}

The Lightweight Optimization Module serves as the core optimization engine of Skills-Coach, designed to continuously enhance both instruction quality and code performance through automated processes. Grounded in Training-Free GRPO\cite{schulman2017proximal,shao2024deepseekmath,cai2025training} and departing from conventional gradient-based parameter optimization, this module leverages the introspective capabilities of LLMs to refine skill instruction and code via a multi-driven mechanism\cite{pryzant2023automatic,yuksekgonul2024textgrad,khattab2023dspy}. This approach enables highly efficient iterative refinement while significantly reducing computational costs and operating entirely autonomously. Notably, the module accelerates training time from hours to minutes, reduces data requirements from thousands of samples to merely dozens, while simultaneously mitigating overfitting risks and demonstrating superior generalization and cross-domain transfer capabilities.

Each optimization epoch comprises two parallel pathways: 1) Instruction Optimization Pathway. Utilizing Training-Free GRPO, the system generates multiple instruction variants, scores them comparatively, and selects the highest-performing variant as the baseline for subsequent iterations, thereby continuously refining instruction quality. 2) Code Optimization Pathway. This employs a three-tier sequential mechanism: a rule-driven optimizer that automatically integrates caching, input validation, and error-handling logic; an LLM-based command optimizer that extracts and refines executable instructions; and an auto-fixer that remediates specific issues—such as dependency conflicts, parameter misconfigurations, and path errors—based on failure case analyses\cite{shinn2023reflexion,madaan2023self,chen2023teaching}. Modifications are executed sequentially by priority, with re-evaluations conducted after each step until convergence or reaching a predefined iteration limit.

To achieve precise and efficient improvements, the module employs differentiated optimization strategies tailored to specific skill categories. For \textit{instruction-only skills}, optimization prioritizes content clarity, structural logic, example sufficiency, and description completeness. For \textit{code-inclusive skills}, refinement extends beyond instruction to encompass code-level improvements, including defect remediation, error-handling augmentation, performance optimization, and overall code quality elevation.

\subsection{Comparative Execution Module}

The Comparative Execution Module serves as the core engine for skill execution and comparison within Skills-Coach. It is responsible for the unbiased execution of both original and optimized skills on identical test tasks, systematically recording comprehensive results for subsequent evaluation. Its primary task is to establish a controlled, isolated, and reproducible execution environment, ensuring that both skill versions operate under strictly identical conditions. By capturing all outputs, side effects, errors, and performance metrics, it provides objective comparative data to the Traceable Evaluation Module. Importantly, this module strictly refrains from performing any scoring or judgment, solely focusing on logging the execution process to guarantee the fairness and accuracy of downstream evaluation.

To achieve these objectives, the module integrates essential operational mechanisms for robust testing. First, the Environment Checker handles pre-execution dependency validation and automated configuration, verifying necessary system commands and provisioning missing dependencies through static analysis of skill specifications. Second, the Skill Executor, as the operational core, provisions an independent temporary workspace for each test task, duplicates its environment, executes commands, and captures standard outputs, errors, return codes, and generated files. Post-execution, it compiles a detailed log and clears the temporary space. Finally, to ensure independence and reproducibility, the module enforces a stringent isolation strategy: tasks run in exclusive temporary directories, original and optimized skills execute sequentially to eliminate order-based dependencies, and temporary directories are immediately purged post-execution to prevent storage depletion.

Furthermore, to maximize operational efficiency, the module incorporates a parallel execution mode, allocating tasks to a thread pool with isolated processing and utilizing thread-safe data structures for metric storage. Concurrently, a Fail-Safe strategy ensures fault tolerance by logging exceptions, capturing error messages, preserving partial outputs, and seamlessly proceeding to the next task upon failure. For complete process traceability, the module generates highly structured outputs, meticulously documenting error details (types, messages, stack traces, system state) in execution logs. The final summary report includes success rate statistics, providing robust empirical support for subsequent failure analysis. Through strict execution isolation, comprehensive performance monitoring, and resilient error handling, this module establishes a fair, objective, and reproducible baseline for skill optimization.

\subsection{Traceable Evaluation Module}

To objectively and quantitatively assess performance differentials between original and optimized skills, we introduce the Traceable Evaluation Module. This module performs multi-dimensional scoring on execution results, computes normalized metrics, conducts comparative analysis, and renders data-driven retention decisions. Its design adheres to five core principles: (1) Scoring Objectivity: scores are deterministically derived from observable execution artifacts, ensuring reproducibility; (2) Criterion Consistency: identical evaluation criteria are applied uniformly to both versions, guaranteeing fair comparison; (3) Analysis Depth: evaluation identifies performance patterns, root causes, and systemic issues beyond surface-level metrics; (4) Decision Rigor: retention decisions are grounded in explicit mathematical rules, eliminating subjective judgment; and (5) Interpretability: comprehensive reports with detailed evidence are generated to ensure full traceability and auditability.

To achieve these objectives, the module comprises four core components. The Criterion Parser extracts evaluation criteria, scoring scales, and passing thresholds from rule documents. The Task Evaluator scores individual task results criterion-by-criterion, supporting both LLM-based deep evaluation and heuristic fallback modes. The Metrics Computer aggregates task-level scores and computes macro-level indicators, including pass rate, average score, standard/advanced task scores, and error rate. The Decision Engine renders retain-or-discard decisions grounded in quantitative results and generates detailed justifications for each verdict.

To maximize evaluation reliability, the module employs a dual-mode strategy. The primary mode leverages LLMs for in-depth assessment across seven dimensions: structural completeness, practicality, example quality, technical depth, clarity, error handling, and comprehensiveness, producing scores (0–100) with detailed supporting evidence\citet{zheng2023judging}. Should an LLM become unavailable or time out, the framework automatically activates an enhanced heuristic mode, which applies multi-dimensional rule-based checks including keyword matching, structural analysis, and content statistics, to ensure continued evaluation robustness.

\section{Experiments}
\label{section: architecture}

\begin{table}[!t]
\centering
\small
\renewcommand{\arraystretch}{1.2} % 增加行高，提升阅读体验
\setlength{\tabcolsep}{1.5mm}     % 稍微放宽列间距
\begin{tabular}{l c c c c c}
\toprule
\textbf{Skill-X}& \textbf{Skills Type}&
\textbf{Average score} & 
\textbf{Pass rate} & 
\textbf{Standard Task score} & 
\textbf{Advanced Task score} \\
\midrule
\multirow{2}{*}{\textbf{All skills}} 
& Original & 0.378 & 33.59\% & 43.00\% & 32.71\% \\
& Optimized & \textbf{0.84} \scriptsize{\textcolor{teal}{(+0.47)}} & \textbf{88.02\%} \scriptsize{\textcolor{teal}{(+54.43\%)}} & \textbf{87.43\%} \scriptsize{\textcolor{teal}{(+44.43\%)}} & \textbf{81.61\%} \scriptsize{\textcolor{teal}{(+48.90\%)}} \\
\midrule
\multirow{2}{*}{\textbf{Instruction-only skills}} 
& Original & 0.388  & 37.93\% & 46.90\% & 33.01\% \\
& Optimized & \textbf{0.839} \scriptsize{\textcolor{teal}{(+0.451)}} & \textbf{91.38\%} \scriptsize{\textcolor{teal}{(+53.45\%)}} & \textbf{89.66\%} \scriptsize{\textcolor{teal}{(+42.76\%)}} & \textbf{79.80\%} \scriptsize{\textcolor{teal}{(+46.78\%)}} \\
\midrule
\multirow{2}{*}{\textbf{Code-inclusive skills}} 
& Original & 0.343  & 26.97\% & 37.06\% & 32.24\% \\
& Optimized & \textbf{0.842} \scriptsize{\textcolor{teal}{(+0.499)}} & \textbf{82.89\%} \scriptsize{\textcolor{teal}{(+55.92\%)}} & \textbf{84.03\%} \scriptsize{\textcolor{teal}{(+46.97\%)}} & \textbf{84.38\%} \scriptsize{\textcolor{teal}{(+52.14\%)}} \\
\bottomrule
\end{tabular}
\caption{Performance Comparison Between Original and Optimized Skills on Skill-X.}
\label{tab:benchmark}
\end{table}

\subsection{Setup}
\textbf{Skill-X.} To comprehensively evaluate the capabilities of Skills-Coach, we introduce Skill-X, a standardized evaluation benchmark encompassing mainstream skills from major developer platforms. Specifically, Skill-X integrates 48 widely used skills sourced from Anthropics, Clawhub, and Vercel Labs. These skills cover a diverse array of real-world application scenarios, ranging from foundational data processing to complex logical interactions (detailed in Appendix \ref{appendix:2}). To ensure a multidimensional assessment, Skill-X categorizes these tools into two primary types: \textit{instruction-only skills}(29 in total) and \textit{code-inclusive skills}(19 in total), thereby achieving comprehensive coverage across varying technology stacks and input modalities. Moving beyond the constraints of isolated, synthetic environments often found in early agent evaluations\cite{liu2023agentbench}, Skill-X systematically evaluates the comprehension, orchestration, execution, and optimization capabilities of Skills-Coach under diverse interface specifications and complex task requirements. Skill-X not only provides multidimensional quantitative metrics but also ensures the universality and real-world applicability of the evaluation results.

\begin{table}[!h]
\centering
\footnotesize
\begin{tabular}{l|c|c|c|c|c}
\toprule
\multirow{2}{*}{\textbf{Skill Name}} & \textbf{Score} & \textbf{Pass Rate} & \textbf{Sta. Task Score} & \textbf{Adv. Task Score} & \multirow{2}{*}{\textbf{Improvement}}\\
& \textbf{Ori} $\to$ \textbf{Opt} & \textbf{Ori} $\to$ \textbf{Opt} & \textbf{Ori} $\to$ \textbf{Opt} & \textbf{Ori} $\to$ \textbf{Opt} & \\
\midrule
\multicolumn{6}{l}{\textit{Perfect Performance (No Improvement Needed)}} \\
\midrule
admapix & 1.0→1.0 & 100\%→100\% & 100\%→100\% & 100\%→100\% & — \\
azure-prepare & 1.0→1.0 & 100\%→100\% & 100\%→100\% & 100\%→100\% & — \\
nano-banana-pro & 1.0→1.0 & 100\%→100\% & 100\%→100\% & 100\%→100\% & — \\
self-improving-agent & 1.0→1.0 & 100\%→100\% & 100\%→100\% & 100\%→100\% & — \\
slack-gif-creator & 1.0→1.0 & 100\%→100\% & 100\%→100\% & 100\%→100\% & — \\
\midrule
\multicolumn{6}{l}{\textit{Exceptional Improvement (Improvement $\geq$ +0.5)}} \\
\midrule
agent-browser-clawdbot & 0.0→0.75 & 0\%→100\% & 0\%→100\% & 0\%→57.1\% & +0.75 \\
azure-ai & 0.17→0.83 & 0\%→100\% & 20\%→100\% & 14.3\%→71.4\% & +0.67 \\
brand-guidelines & 0.17→0.83 & 0\%→100\% & 20\%→80\% & 14.3\%→85.7\% & +0.67 \\
browser & 0.0→1.0 & 0\%→100\% & 0\%→100\% & 0\%→100\% & +1.0 \\
composition-patterns & 0.17→1.0 & 0\%→100\% & 20\%→100\% & 14.3\%→100\% & +0.83 \\
find-skills & 0.17→1.0 & 0\%→100\% & 20\%→100\% & 14.3\%→100\% & +0.83 \\
internal-comms & 0.17→0.67 & 0\%→100\% & 20\%→80\% & 14.3\%→57.1\% & +0.5 \\
mcp-builder & 0.0→1.0 & 0\%→100\% & 0\%→100\% & 0\%→100\% & +1.0 \\
ontology & 0.0→1.0 & 0\%→100\% & 0\%→100\% & 0\%→100\% & +1.0 \\
polymarket-trade & 0.21→1.0 & 12.5\%→100\% & 33.3\%→100\% & 12.5\%→100\% & +0.79 \\
react-best-practices & 0.5→1.0 & 50\%→100\% & 60\%→100\% & 42.9\%→100\% & +0.5 \\
react-native-skills & 0.33→1.0 & 50\%→100\% & 60\%→100\% & 14.3\%→100\% & +0.67 \\
remotion-best-practices & 0.33→1.0 & 50\%→100\% & 60\%→100\% & 14.3\%→100\% & +0.67 \\
rss-daily-digest & 0.0→1.0 & 0\%→100\% & 0\%→100\% & 0\%→100\% & +1.0 \\
self-evolving-skill & 0.14→0.82 & 0\%→75\% & 16.7\%→83.3\% & 12.5\%→81.2\% & +0.68 \\
skill-creator & 0.14→0.66 & 0\%→75\% & 16.7\%→70.8\% & 12.5\%→62.5\% & +0.52 \\
stock-analysis & 0.14→1.0 & 0\%→100\% & 16.7\%→100\% & 12.5\%→100\% & +0.86 \\
theme-factory & 0.33→1.0 & 50\%→100\% & 60\%→100\% & 14.3\%→100\% & +0.67 \\
ui-ux-pro-max & 0.21→1.0 & 0\%→100\% & 25\%→100\% & 18.8\%→100\% & +0.79 \\
vercel-react-best-practices & 0.5→1.0 & 50\%→100\% & 60\%→100\% & 42.9\%→100\% & +0.5 \\
weather & 0.5→1.0 & 50\%→100\% & 80\%→100\% & 28.6\%→100\% & +0.5 \\
web-design-guidelines & 0.33→0.83 & 50\%→100\% & 60\%→80\% & 14.3\%→85.7\% & +0.5 \\
xlsx & 0.36→1.0 & 0\%→100\% & 50\%→100\% & 25\%→100\% & +0.64 \\
\midrule
\multicolumn{6}{l}{\textit{Significant Improvement (+0.3 to +0.49)}} \\
\midrule
algorithmic-art & 0.33→0.75 & 0\%→100\% & 20\%→80\% & 42.9\%→71.4\% & +0.42 \\
azure-deploy & 0.25→0.67 & 0\%→100\% & 20\%→80\% & 28.6\%→57.1\% & +0.42 \\
byterover & 0.67→1.0 & 100\%→100\% & 80\%→100\% & 57.1\%→100\% & +0.33 \\
doc-coauthoring & 0.25→0.58 & 0\%→50\% & 20\%→80\% & 28.6\%→42.9\% & +0.33 \\
docx & 0.0→0.36 & 0\%→37.5\% & 0\%→37.5\% & 0\%→35\% & +0.36 \\
microsoft-foundry & 0.5→0.92 & 50\%→100\% & 40\%→100\% & 57.1\%→85.7\% & +0.42 \\
multi-search-engine & 0.25→0.67 & 0\%→100\% & 20\%→80\% & 28.6\%→57.1\% & +0.42 \\
pollyreach & 0.52→0.88 & 50\%→100\% & 70.8\%→70.8\% & 37.5\%→100\% & +0.36 \\
self-improving-proactive-agent & 0.33→0.75 & 50\%→100\% & 60\%→80\% & 14.3\%→71.4\% & +0.42 \\
vercel-cli-with-tokens & 0.67→1.0 & 100\%→100\% & 80\%→100\% & 57.1\%→100\% & +0.33 \\
web-artifacts-builder & 0.67→1.0 & 100\%→100\% & 80\%→100\% & 57.1\%→100\% & +0.33 \\
webapp-testing & 0.07→0.5 & 0\%→0\% & 0\%→50\% & 12.5\%→50\% & +0.43 \\
\midrule
\multicolumn{6}{l}{\textit{Moderate Improvement (+0.1 to +0.29)}} \\
\midrule
azure-diagnostics & 0.5→0.67 & 50\%→100\% & 60\%→80\% & 42.9\%→57.1\% & +0.17 \\
canvas-design & 0.17→0.42 & 0\%→50\% & 20\%→60\% & 14.3\%→28.6\% & +0.25 \\
deploy-to-vercel & 0.75→1.0 & 100\%→100\% & 80\%→100\% & 71.4\%→100\% & +0.25 \\
frontend-design & 0.17→0.42 & 0\%→0\% & 20\%→40\% & 14.3\%→42.9\% & +0.25 \\
obsidian-1-0-0 & 0.17→0.33 & 0\%→50\% & 20\%→60\% & 14.3\%→14.3\% & +0.17 \\
pdf & 0.43→0.54 & 50\%→50\% & 41.7\%→54.2\% & 43.8\%→53.1\% & +0.11 \\
pptx & 0.29→0.5 & 0\%→37.5\% & 33.3\%→50\% & 25\%→50\% & +0.21 \\
\midrule
\multicolumn{6}{l}{\textit{Limited Improvement (below +0.1)}} \\
\midrule
react-view-transitions & 0.92→1.0 & 100\%→100\% & 100\%→100\% & 85.7\%→100\% & +0.08 \\
\bottomrule
\end{tabular}
\caption{Skill Performance Analysis: Original vs Optimized Results (48 Skills) organized by Improvement Level.}
\label{table:skill_comparison_optimized}
\end{table}

\textbf{Metrics.} We quantify performance by strictly evaluating each task's execution result against predefined itemized criteria, awarding one point for every satisfied condition. A task's final score is the cumulative sum of these points. Furthermore, a task is deemed "passed" if its score meets or exceeds a predefined threshold (defaulting to 70\% of the maximum possible score). Based on this scoring mechanism, our overall assessment encompasses the pass rate, average score, standard task score, and advanced task score.

The specific criteria vary based on the task type. For instruction-only skills, LLMs are employed to conduct in-depth evaluations across eight dimensions: structural integrity, practicality, example quality, technical depth, clarity, command coverage, error handling, and advanced scenario adaptability. For code-inclusive skills, the assessment focuses on deterministic outcomes, including error-free command execution, successful output generation, format validity, expected data matching, and the robustness of handling complex scenarios.

\textbf{Experimental Setting.}
For each target skill, Skills-Coach runs three optimization epochs by default (num\_epochs=3).
In task generation, the current pipeline creates 12 training tasks and 8 test tasks, with an even split between difficulty levels (6 standard + 6 advanced for training, 4 standard + 4 advanced for testing).
During Training-Free GRPO optimization, each epoch generates 3 rollout variants (group\_size=3) for the current skill version and evaluates them on the training set, rather than generating variants per individual task.
After execution and failure analysis, the Auto-Fix loop supports up to 2 iterative fix-test-reanalyze rounds (max\_iterations=2).
Execution defaults to real mode for both optimization-time execution and final evaluation.
For LLM-powered optimization/fixing components, the default model is claude-sonnet-4-6 (with heuristic fallbacks when API-based generation is unavailable).

\subsection{Main Result}

We evaluate the performance of Skills-Coach on the Skill-X. As presented in Table \ref{tab:benchmark}, Skills-Coach demonstrates significant performance improvements across all evaluation metrics. Specifically, the average score markedly increases from 0.37 to 0.84, representing a substantial 127\% relative growth. Concurrently, the pass rate dramatically improves from 33.59\% to 88.02\%, indicating a robust increase of 54.43\%. These compelling results underscore Skills-Coach's effectiveness in enhancing both skill generalization and practical application capabilities through its systematic optimization strategies.

\begin{figure*}[!t] 
    \centering 
    \includegraphics[width=1\textwidth]{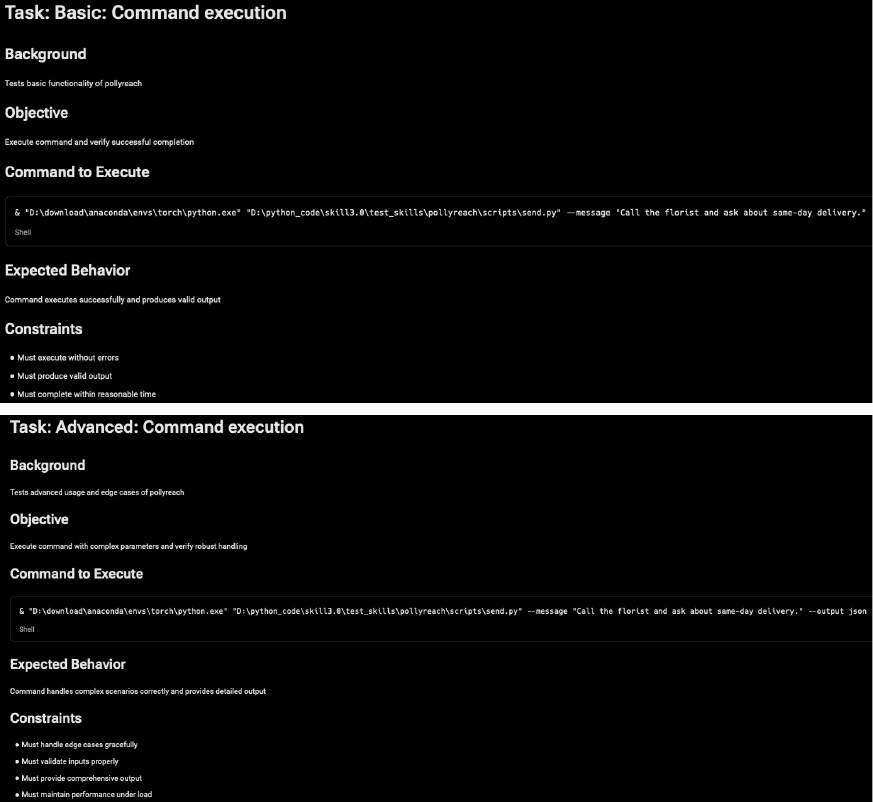}
    \caption{Comparison of Standard and Advanced Test Tasks Generated for Pollyreach.}
    \label{fig:case1} 
\end{figure*}

\begin{figure*}[!htbp] 
    \centering 
    \includegraphics[width=1\textwidth]{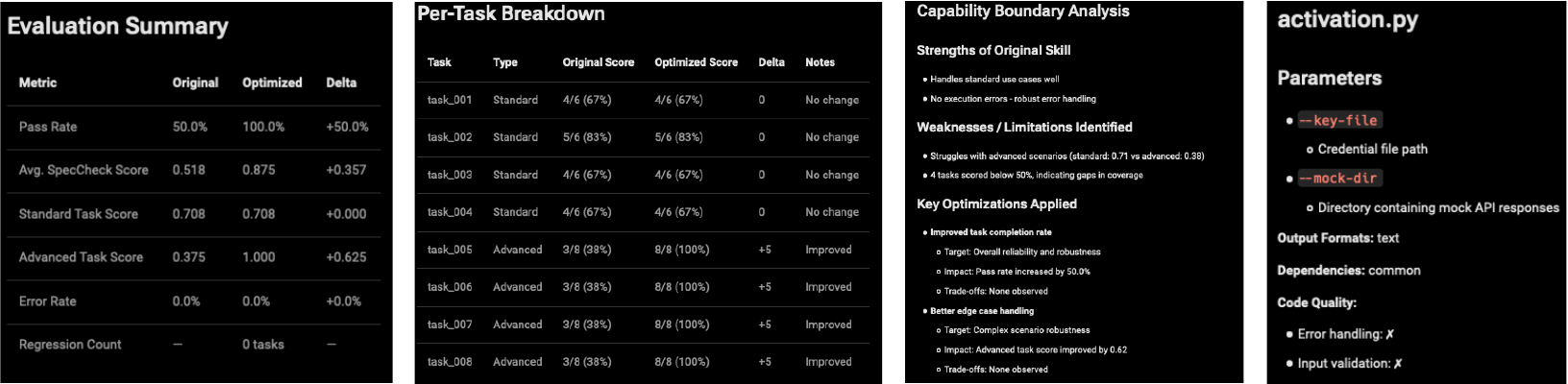}
    \caption{Key Contents from the Pollyreach Summary Report.}
    \label{fig:case2} 
\end{figure*}

Furthermore, an individual examination of different skill types reveals consistent improvement trends. Both instruction-only skills and code-inclusive skills achieve over a 50\% increase in pass rate. Notably, Skills-Coach exhibits the largest relative improvement for code-inclusive skills, suggesting its particular efficacy for tasks demanding complex logical reasoning. Consistent performance gains are also observed across task difficulties: while both standard and advanced tasks show improved performance, advanced tasks demonstrate an even greater magnitude of improvement than standard tasks, highlighting Skills-Coach's distinct advantage in tackling more challenging problems.

\subsection{Further Analyses}

Table \ref{table:skill_comparison_optimized} systematically evaluates the optimization results for 48 skills, demonstrating the significant effectiveness of the optimization strategy. In the \textit{Exceptional Improvement} category, 23 skills showed score increases of +0.5 or higher. Among these, four skills, including ``Browser'' and ``Ontology,'' experienced dramatic improvements, with scores leaping from 0.0 to 1.0 (an improvement of +1.0) and pass rates rising from 0\% or lower levels to 100\% or nearly 100\%. These results indicate that Skills-Coach can deliver transformative improvements to underperforming skills, elevating them to or near optimal performance levels.

However, as the intrinsic performance of skills improves, the marginal benefit of optimization exhibits diminishing returns. In the \textit{Perfect Performance} category, five skills maintained perfect scores of 1.0 and 100\% pass rates both before and after optimization, suggesting they had already achieved optimal status in their original form and required no further improvement. Conversely, in the \textit{Limited Improvement} category, only one skill, ``react-view-transitions,'' showed score improvement from 0.92 to 1.0, a modest gain of +0.08. This phenomenon underscores the critical importance of strategically allocating optimization resources: concentrating efforts on skills with low intrinsic performance and substantial room for improvement (such as those in the \textit{Exceptional Improvement} and \textit{Significant Improvement} categories) yields the highest return on investment. In contrast, additional optimization for skills already approaching perfection yields negligible returns. These findings provide valuable quantitative evidence for prioritizing future skill development and optimization initiatives.

\subsection{Case Study}

Figure \ref{fig:case1} illustrates a comparison of basic and advanced test tasks designed for Pollyreach. Basic tasks assess the standard execution capability of Pollyreach commands, requiring the system to correctly parse command paths and generate valid outputs. Advanced tasks elevate complexity by incorporating same-day delivery functionalities, demanding that the system produce results in JSON format and rigorously validate input parameters. This hierarchical design embodies a progressive testing strategy that advances from functional correctness to robustness and edge-case handling.

Figure \ref{fig:case2} presents key aspects of the Pollyreach optimization summary report. This report encompasses two principal dimensions: Per-Task Breakdown and Capability Boundary Analysis. It systematically illustrates comparative improvements between original and optimized skills across multiple metrics, including pass rate, average score, standard task score, and advanced task score. The structured report format provides comprehensive visualization of these metrics, while the multi-dimensional data comparison supplies robust quantitative support for subsequent iterative optimization cycles.

\section{Conclusion}
\label{section: conclusion}

In this work, we introduce Skills-Coach, a noval automated framework designed to empower LLM-based agents with enhanced self-evolution capabilities for their skills. By addressing the critical challenge of a fragmented skill ecosystem, Skills-Coach paves the way for more comprehensive and robust intelligent applications. Our framework makes several key contributions. First, we formalize the skill self-evolution problem and propose an automated solution that requires no human intervention, significantly reducing the burden on skill developers and maintainers. Second, we demonstrate that Training-Free GRPO is an efficient alternative to conventional parameter optimization, achieving competitive results in minutes rather than hours while dramatically reducing data requirements. Third, through rigorous evaluation on Skill-X with 48 diverse skills, we conclusively demonstrate substantial performance improvements across diverse skill categories and task complexities, with particularly pronounced gains for code-inclusive skills and advanced tasks. By enabling autonomous skill optimization at scale,  Skills-Coach has the potential to transform how developers interact with and leverage skill ecosystems, ultimately advancing the practical deployment of intelligent applications across diverse industrial domains.
% \clearpage
% \newpage
% \bibliographystyle{assets/plainnat}
\bibliographystyle{unsrtnat}

\bibliography{ref}

\clearpage
\newpage
\beginappendix

\section{Evaluation Dimensions with 51 Discrete Evaluation Metrics}
\label{appendix:1}

As shown in Table \ref{tab:evaluation_standards}, the generated samples encompass eight evaluation dimensions, comprising a total of 51 independent evaluation metrics. This ensures that the generated samples are not only functional but also meet professional-grade standards suitable for production environments.

\begin{table}[h!]
\centering
\scriptsize
\begin{tabular}{p{2.5cm}|p{12.5cm}}
\toprule
\textbf{Dimension} & \textbf{Assessment Criteria} \\
\midrule
\textbf{1. Structural Completeness \& Organization (7 points)} & 
\begin{minipage}[t]{\linewidth}
\begin{enumerate}
\item Clear introduction/overview at document start explaining purpose and goals
\item Installation/setup instructions with complete environment configuration
\item Comprehensive usage section detailing all commands and functions
\item Multiple concrete examples with at least 3 different real-world scenarios
\item Configuration/parameters section listing all configurable options
\item Troubleshooting/error handling with dedicated section for FAQs
\item Logical progression from basic to advanced concepts
\end{enumerate}
\end{minipage} \\
\midrule

\textbf{2. Practical Usability \& Learnability (6 points)} & 
\begin{minipage}[t]{\linewidth}
\begin{enumerate}
\item Beginner step-by-step guide with clear guidance keywords (first, then, next)
\item Copy-paste ready examples with actual commands (\$, python, bash, etc.)
\item Explicit prerequisites clearly listing dependencies and required knowledge
\item Common pitfalls documentation with warning/note/important markers
\item Progressive complexity from simple to advanced examples
\item Quick start guide or minimal working example section
\end{enumerate}
\end{minipage} \\
\midrule

\textbf{3. Example Quality \& Coverage (6 points)} & 
\begin{minipage}[t]{\linewidth}
\begin{enumerate}
\item At least 3 different real examples with complete executable code blocks
\item Diverse use cases covering different scenarios, not just task variations
\item Expected output demonstration using output:/result:/=>/-> markers
\item Boundary condition examples showing edge cases and extreme scenarios
\item Error handling scenarios demonstrating exception and failure handling
\item Complex multi-step workflow showing complete real-world application
\end{enumerate}
\end{minipage} \\
\midrule

\textbf{4. Technical Depth \& Accuracy (6 points)} & 
\begin{minipage}[t]{\linewidth}
\begin{enumerate}
\item All parameters/options documented with parameter/option/flag keywords
\item Return values and output format specification (types, JSON structure)
\item Performance characteristics mentioned when relevant
\item Clear limitations and constraints explicitly listed
\item Integration with other systems explained and demonstrated
\item Correct use of 2+ professional technical terms (API, CLI, SDK, etc.)
\end{enumerate}
\end{minipage} \\
\midrule

\textbf{5. Clarity \& Readability (6 points)} & 
\begin{minipage}[t]{\linewidth}
\begin{enumerate}
\item Clear concise language with average sentence length < 30 words
\item Consistent formatting and style with unified header levels
\item Proper use of at least 3 headers, lists (- or *), and code blocks
\item Unambiguous statements avoiding vague or misleading expressions
\item Appropriate detail level (500-15000 characters, not too brief or verbose)
\item Good visual hierarchy using secondary headers (\texttt{\textbackslash\#\#}) or tertiary headers (\texttt{\textbackslash\#\#\#})
\end{enumerate}
\end{minipage} \\
\midrule

\textbf{6. Command Coverage Completeness (6 points)} & 
\begin{minipage}[t]{\linewidth}
\begin{enumerate}
\item Every command in examples explained in documentation
\item All flags/options for each command documented
\item Command syntax clearly demonstrated with correct format
\item Usage context explained for when to use each command
\item Relationships between multiple commands clarified
\item No undocumented or hidden functionality
\end{enumerate}
\end{minipage} \\
\midrule

\textbf{7. Error Handling \& Troubleshooting (6 points)} & 
\begin{minipage}[t]{\linewidth}
\begin{enumerate}
\item Common errors and solutions listed with fixes
\item Error message explanations clarifying meaning and context
\item Debugging tips provided with diagnostic methods and commands
\item Known issues and workarounds documented
\item Support and bug reporting instructions provided
\item Verification steps to check configuration correctness
\end{enumerate}
\end{minipage} \\
\midrule

\textbf{8. Advanced Scenarios \& Best Practices (6 points)} & 
\begin{minipage}[t]{\linewidth}
\begin{enumerate}
\item Advanced use cases and patterns with advanced/complex/production examples
\item Best practices and recommendations using best practice/recommended/tip keywords
\item Performance optimization tips when applicable
\item Security considerations mentioned and explained when relevant
\item Integration patterns showing how to combine with other tools
\item Real-world workflow examples demonstrating complete practical scenarios
\end{enumerate}
\end{minipage} \\
\bottomrule
\end{tabular}
\caption{8 Evaluation Dimensions with 51 Specific Assessment Criteria}
\label{tab:evaluation_standards}
\end{table}

\section{Skill Sources in Skill-X}
\label{appendix:2}

\begin{tabular}{lll}
\toprule
skill name & source & rank in source \\
\midrule
self-improving-agent & clawhub.ai & 1 \\
ontology & clawhub.ai & 3 \\
self-improving-proactive-agent & clawhub.ai & 4 \\
weather & clawhub.ai & 8 \\
multi-search-engine & clawhub.ai & 9 \\
polymarket-trade & clawhub.ai & 10 \\
pollyreach & clawhub.ai & 11 \\
admapix & clawhub.ai & Common \\
agent-browser-clawdbot & clawhub.ai & Common \\
byterover & clawhub.ai & Common \\
nano-banana-pro & clawhub.ai & Common \\
obsidian-1-0-0 & clawhub.ai & Common \\
rss-daily-digest & clawhub.ai & Common \\
self-evolving-skill & clawhub.ai & Common \\
stock-analysis & clawhub.ai & Common \\
algorithmic-art & github.com/anthropics/skills & Common \\
brand-guidelines & github.com/anthropics/skills & Common \\
canvas-design & github.com/anthropics/skills & Common \\
doc-coauthoring & github.com/anthropics/skills & Common \\
docx & github.com/anthropics/skills & Common \\
frontend-design & github.com/anthropics/skills & Common \\
internal-comms & github.com/anthropics/skills & Common \\
mcp-builder & github.com/anthropics/skills & Common \\
pdf & github.com/anthropics/skills & Common \\
pptx & github.com/anthropics/skills & Common \\
skill-creator & github.com/anthropics/skills & Common \\
slack-gif-creator & github.com/anthropics/skills & Common \\
theme-factory & github.com/anthropics/skills & Common \\
web-artifacts-builder & github.com/anthropics/skills & Common \\
webapp-testing & github.com/anthropics/skills & Common \\
xlsx & github.com/anthropics/skills & Common \\
composition-patterns & github.com/vercel-labs/agent-skills/tree/main/skills & Common \\
deploy-to-vercel & github.com/vercel-labs/agent-skills/tree/main/skills & Common \\
react-best-practices & github.com/vercel-labs/agent-skills/tree/main/skills & Common \\
react-native-skills & github.com/vercel-labs/agent-skills/tree/main/skills & Common \\
react-view-transitions & github.com/vercel-labs/agent-skills/tree/main/skills & Common \\
vercel-cli-with-tokens & github.com/vercel-labs/agent-skills/tree/main/skills & Common \\
vercel-react-best-practices & github.com/vercel-labs/agent-skills/tree/main/skills & Common \\
web-design-guidelines & github.com/vercel-labs/agent-skills/tree/main/skills & Common \\
find-skills & skills.sh & 1 \\
remotion-best-practices & skills.sh & 5 \\
microsoft-foundry & skills.sh & 6 \\
azure-ai & skills.sh & 7 \\
azure-deploy & skills.sh & 8 \\
azure-prepare & skills.sh & 9 \\
azure-diagnostics & skills.sh & 10 \\
browser & skills.sh & Common \\
ui-ux-pro-max & skills.sh & Common \\
\bottomrule
\end{tabular}

\end{document}